\documentclass{article}
\usepackage{ijcai22}

\usepackage{times}
\usepackage{soul}
\usepackage{url}
\usepackage[hidelinks]{hyperref}
\usepackage[utf8]{inputenc}
\usepackage[small]{caption}
\usepackage{graphicx}
\usepackage{amsmath}
\usepackage{amsthm}
\usepackage{booktabs}
\usepackage{algorithm}
\usepackage{algorithmic}
\usepackage{comment}
\usepackage{amsmath}
\usepackage{amssymb}
\usepackage{mathrsfs}
\usepackage{booktabs}
\usepackage{mathtools}
\usepackage{pifont}
\usepackage{color}
\usepackage{diagbox}
\usepackage{enumerate}
\usepackage{array}
\usepackage{multirow}
\usepackage{epstopdf}
\usepackage{url}
\usepackage{algorithm}
\usepackage{algorithmic}
\usepackage{subfigure}
\usepackage{bbding}

\newcommand\ourmethod{UM4}

\newcommand{\cmark}{\ding{51}}

\title{\ourmethod{}: Unified Multilingual Multiple Teacher-Student Model for \\ Zero-Resource Neural Machine Translation}

\author{}

\author{
  Jian Yang\textsuperscript{\rm 1 \thanks{\ Equal contribution. Work done during internship at Microsoft.}}, 
  Yuwei Yin\textsuperscript{\rm 2 *}, 
  Shuming Ma\textsuperscript{\rm 2}, 
  Dongdong Zhang\textsuperscript{\rm 2}, 
  Shuangzhi Wu\textsuperscript{\rm 3}, \\
  Hongcheng Guo\textsuperscript{\rm 2}, 
  Zhoujun Li\textsuperscript{\rm 1 \thanks{\ Corresponding author.}}, 
  Furu Wei\textsuperscript{\rm 2} 
  \affiliations
  $^1$State Key Lab of Software Development Environment, Beihang University
  \\ 
  $^2$Microsoft Research\\
  $^3$Tencent Cloud Xiaowei \\
  \{jiaya, lizj\}@buaa.edu.cn,  frostwu@tencent.com, \\ \{v-yuweiyin, shumma, dozhang, v-hongguo, fuwei\}@microsoft.com
}

\begin{document}

\maketitle

\begin{abstract}
Most translation tasks among languages belong to the zero-resource translation problem where parallel corpora are unavailable. Multilingual neural machine translation (MNMT) enables one-pass translation using shared semantic space for all languages compared to the two-pass pivot translation but often underperforms the pivot-based method. In this paper, we propose a novel method, named as \textbf{U}nified \textbf{M}ultilingual \textbf{M}ultiple teacher-student \textbf{M}odel for N\textbf{M}T (\textbf{\ourmethod{}}). Our method unifies source-teacher, target-teacher, and pivot-teacher models to guide the student model for the zero-resource translation. The source teacher and target teacher force the student to learn the direct source$\to$target translation by the distilled knowledge on both source and target sides. The monolingual corpus is further leveraged by the pivot-teacher model to enhance the student model. Experimental results demonstrate that our model of 72 directions significantly outperforms previous methods on the WMT benchmark. 
\end{abstract}

\section{Introduction}
The encoder-decoder framework \cite{transformer,cross_attention_pretrained_transformer} has gained outstanding performance on rich-resource machine translation tasks, such as English-German, English-French, and Chinese-English \cite{WMT2019,robust_mnmt,multilingual}, where large-scale parallel corpora are available. However, it is incapable of directly modeling the zero-resource translation task when the parallel training data does not exist.

\begin{figure}[t]
\begin{center}
    \includegraphics[width=1.0\columnwidth]{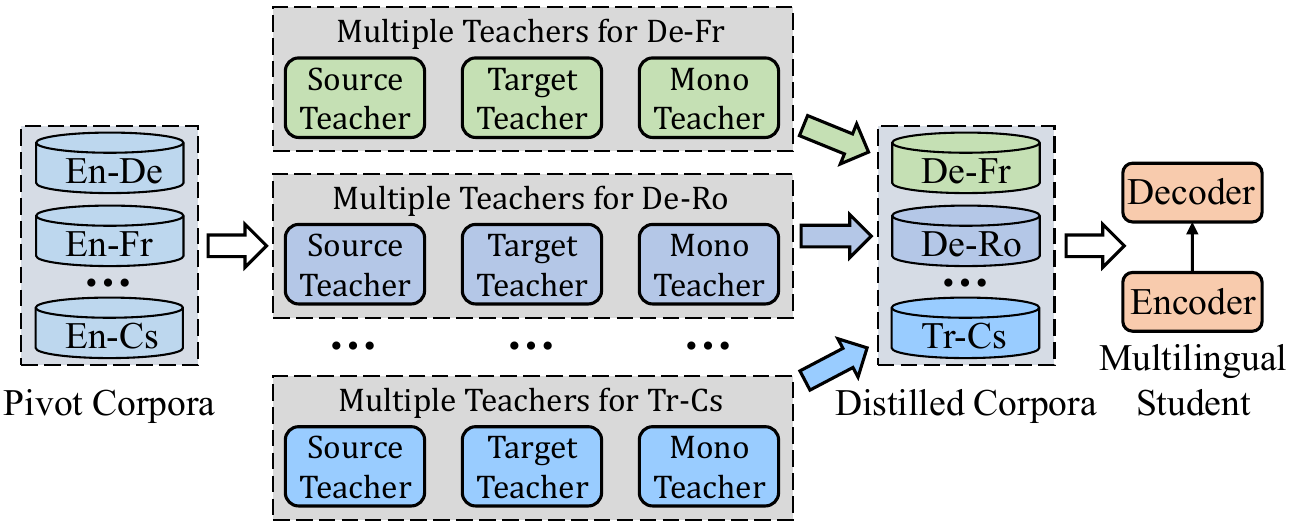}
    \caption{Framework of \ourmethod{}. Unified multiple teachers with shared parameters trained on the original pivot corpora are used to guide the multilingual student model. English (En) is the pivot language.}
    \label{intro}
    \vspace{-10pt}
\end{center}
\end{figure}

A straightforward solution for the zero-resource machine translation problem is the pivot translation approach \cite{SMT_pivot1,SMT_pivot2,SMT_pivot3,joint_training}. Bilingual pivot-based models perform the two-pass translation, which increases the computation cost and potentially suffers from the error propagation problem \cite{error_propagation}. There are also some works \cite{syntheticdata_teacher_student,zero_resource_pivot_mono,pivot-transfer-learning} directly building the source$\to$target model but limited by the bilingual translation task. Beyond pivot-based methods, the multilingual model trained over multiple pivot corpora with shared parameters utilizes the language symbol to infer the desired translation direction \cite{cascade_5,multilingual,multilingual_pivot,zero_resource_pivot_mono}. The multilingual model benefits from different language pairs and only requires one-step translation, which avoids error propagation and saves inference time.
But the performance of this approach \cite{pivot_based_transfer_learning} is worse than pivot-based models. 

\begin{figure*}[t]
\centering
    \subfigure[]{
    \includegraphics[width=0.52\columnwidth]{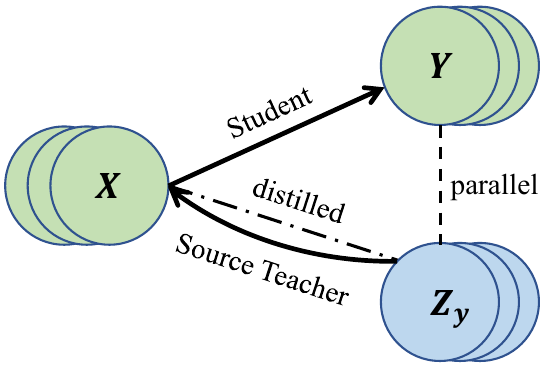}\quad
    \label{source-teacher}
    }
    \subfigure[]{
    \includegraphics[width=0.52\columnwidth]{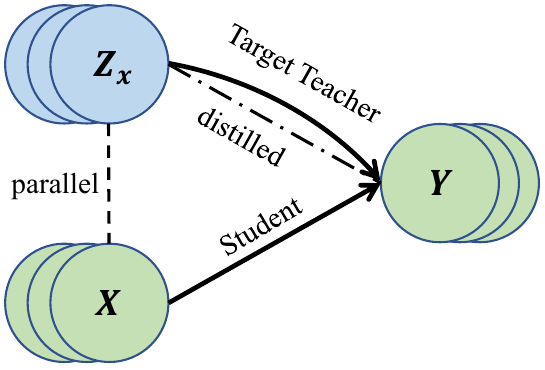}\quad
    \label{teacher-student}
    }
    \subfigure[]{
    \includegraphics[width=0.52\columnwidth]{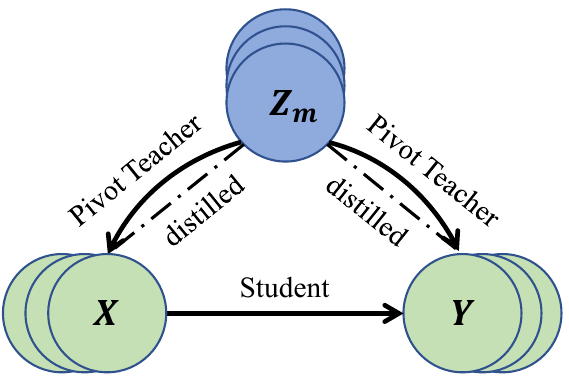}
    \label{pivot-teacher}
    }
    \caption{Overview of our unified multilingual multiple teacher-student model: (a) the source-teacher model, (b) target-teacher model, and (c) the pivot-teacher model. $X,Y,Z$ separately denote the source, target, and pivot language. The dashed line ``- -'' denotes a real parallel corpus is available between the connected language pair and the dotted line ``$\cdot$ $\cdot$'' denotes a distilled parallel corpus generated by the teacher model is available. Solid arrow lines represent translation directions. 
    Our multiple teacher models consist of a source-teacher model, a target-teacher, and a pivot-teacher model, where $Z_x$ and $Z_y$ represent pivot language $Z$ in the parallel corpus associated with source language $X$ and target language $Y$ respectively. The source teacher and target teacher transfers the knowledge from $Z_y$ and $Z_x$ separately.  
    Given the monolingual corpus $Z_{m}$, the pivot-teacher model further enhances the multilingual student model by distilling knowledge to both source and target sides.}
    \label{overview}
\end{figure*}

Along the line of leveraging the multilingual model to address the zero-resource translation problem, we propose a novel method called \textbf{U}nified \textbf{M}ultilingual \textbf{M}ultiple teacher-student \textbf{M}odel for N\textbf{M}T (\textbf{\ourmethod{}}). Given the available corpora of the pivot and other languages, we directly build the source$\to$target student translation model guided by the multilingual multiple teachers as shown in Figure \ref{intro}. The multiple teacher models can be decomposed into a source-teacher model, a target-teacher model, and a pivot-teacher model. The source-teacher model transfer the knowledge from the pivot to the source sentence. The target teacher distills pivot knowledge to the target side and boosts the capability of the target generation. The pivot-teacher model further enhances the student model by mining the potential of monolingual pivot corpora.
The overall distilled corpora from unified teachers with normalized scores are used for the student.

Specifically, we first train a single multilingual model on all pivot corpora as unified multiple teachers sharing all model parameters. Then, we construct the distilled multilingual corpora of all zero-resource directions using multiple teacher models. The distilled corpora with the normalized scores generated by the unified multiple teachers are used to guide a source$\to$target student. We conduct experiments on the multilingual corpora from the WMT benchmark of 9 languages with 72 translation directions. The experimental results show that our method can significantly outperform multilingual baselines and pivot-based methods. Furthermore, we verify the effectiveness of our method by perturbation experiments and visualization of the multilingual sentence representations. Analytic results demonstrate that our \ourmethod{} student model with better crosslingual ability enhances zero-resource translations and avoids error propagation.

\section{Our Approach}

In this section, we introduce the unified multiple teacher-student model for zero-resource machine translation. As illustrated in Figure \ref{overview}, our method simultaneously uses multiple teacher models to train a multilingual end-to-end translation model when the direct parallel data is unavailable.

\subsection{Overview of \ourmethod{}}
\label{Multilingual Machine Translation}
Given the bilingual corpora $D_{B}=\{D_{B_n}\}_{n=1}^{N}$ of $N$ languages, where one side is the pivot language $L_z$ and the other side is the language $L_n \in \{L_n\}_{n=1}^{N}$, the multilingual model is trained on the available pivot corpora $D_B$ to solve the zero-resource translation without direct parallel data between the zero-resource pair $L_i$ and $L_j$ ($1 \leq i,j \leq N$ and $i \neq j$):
\begin{BigEquation}
\begin{align}
\begin{split}
\mathcal{L}_{D} &=\sum_{n=1}^{N} \mathbb{E}_{x,z_x \in D_{B_n}} \left[ -\log P_{\theta}(z_x|x) \right] \\
                &+\sum_{n=1}^{N} \mathbb{E}_{y,z_y \in D_{B_n}} \left[ -\log P_{\theta}(y|z_y) \right] 
    \label{objective-d}
\end{split}
\end{align}
\end{BigEquation}where $x,z_x$ denote the source and pivot language sentence in the bilingual corpus $D_{B}$. $y,z_y$ denote the pivot and target sentence.
$\mathcal{L}_{D}$ is the combined objective of the multilingual model. The multilingual model over the source-pivot and pivot-target corpora with shared parameters prepends the language symbol to indicate the zero-resource translation direction from language $L_i$ to language $L_j$. 

Without parallel training data for zero-resource language pairs, the multilingual model can easily translate into a wrong language and result in poor translation quality. Therefore, we introduce the synthetic multilingual multiple corpora of zero-resource language pairs $D_{S}=\{D_{S_m}\}_{m=1}^{M}$.
\begin{BigEquation}
\begin{align}
    \begin{split}
    \mathcal{L}_{S} &=\sum_{m=1}^{M} \mathbb{E}_{x,y \in D_{S_m}} \left[ -w_{x,y} \log P_{\theta}(y|x) \right] 
    \label{objective-s}
\end{split}
\end{align}
\end{BigEquation}where $x$ and $y$ denote the source and target language sentence in the distilled multilingual corpora $D_{S}$. $w_{x,y}$ is the weight of the training sample from multilingual multiple teachers. 

Our multilingual student model is trained on both original corpora $D_B$ and distilled corpora $D_S$, which improves the translation quality under the supervised manner among zero-resource directions: 
\begin{BigEquation}
\begin{align}
    \begin{split}
 \mathcal{L}_{T} = \mathcal{L}_{D} + \mathcal{L}_{S}
 \label{total_training_objective}
\end{split}
\end{align}
\end{BigEquation}where $\mathcal{L}_{T}$ is the total objective of our multilingual student model. $\mathcal{L}_{D}$ and $\mathcal{L}_{S}$ denote training objective over the original pivot corpora $D_{B}$ and distilled corpora $D_{S}$ respectively.

\subsection{Multiple Teacher Models}
\label{Coupled Teacher-Student Model}
Formally, given the source-pivot and target-pivot parallel corpus $D_{B_i}=\{x^{(k)},z_{x}^{(k)}\}^{\lvert D_{B_i} \rvert}_{k=1}$ and $D_{B_j}=\{y^{(k)},z_{y}^{(k)}\}^{\lvert D_{B_j} \rvert}_{k=1}$, we aim to build a source$\to$target translation model $\theta_{x \to y}$ for the zero-resource translation task.
$x$ and $y$ denote the source and target sentence respectively, $z_x$ and $z_y$ denote the pivot sentence from the source-pivot corpus $D_{B_{i}}$ and pivot-target corpus $D_{B_{j}}$ separately.
$\lvert D_{B_i} \rvert$ and $\lvert D_{B_j} \rvert$ are the size of corpora $D_{B_{i}}$ and $D_{B_{j}}$. $\theta$ denote model parameters.

\paragraph{Source-teacher Model} If the target sentence $y$ and pivot sentence $z_{y}$ are parallel from the dataset $D_{B_{j}}$. The source teacher $\theta_{z_{y} \to x}$ is trained on the source-pivot corpus $D_{B_{i}}$. The source-teacher student training objective can be written as:
\begin{BigEquation}
\begin{align}
    \mathcal{L}_{S}^{src} \! &= -\mathbb{E}_{y,z_y \in D_{B_m}} \left[ P(x|z_{y};\theta_{z_y \to x}) \log P_{\theta}(y|x) \right] 
    \label{source-teacher-training}
\end{align}
\end{BigEquation}where $P(x|z_{y};\theta_{z_{y} \to x})$ is the weight generated by the source-teacher model $\theta_{z_{y} \to x}$.

\paragraph{Target-teacher Model} 
If the source sentence $x$ and pivot sentence $z_{x}$ are parallel from the dataset $D_{B_{i}}$. The target teacher $\theta_{z_x \to y}$ on the pivot-target corpus $D_{B_{j}}$. The target-teacher student training objective can described as:
\begin{BigEquation}
\begin{align}
     \mathcal{L}_{S}^{tgt} &= -\mathbb{E}_{x,z_x \in D_{B_i}} \left[ P(y|z_x;\theta_{z_x \to y}) \log P_{\theta}(y|x) \right] 
     \label{target-teacher-training}
\end{align}
\end{BigEquation}where $P(y|z_{x};\theta_{z_x \to y})$ is the weight generated by the target-teacher model $\theta_{z_x \to y}$.

\paragraph{Pivot-teacher Model}
Given the monolingual pivot corpus $D_{M}$, the pivot-teacher is used to guide the student model. The pivot-teacher model $\theta_{z_m \to x} \cup \theta_{z_m \to y}$ is trained on the pivot corpora $D_{B_{i}}$ and $D_{B_{j}}$. The pivot-teacher student training objective can described as:
\begin{BigEquation}
\begin{align}
     \mathcal{L}_{S}^{pivot} &= -\mathbb{E}_{z_{m} \in D_{M}} \left[ w_{x,y} \log P_{\theta}(y|x) \right] 
     \label{pivot-teacher-training}
\end{align}
\end{BigEquation}where $w_{x,y}=P(y|z_{m};\theta_{z_{m} \to y})P(x|z_{m};\theta_{z_{m} \to x})$ is the weight generated by the pivot-teacher model.

All teachers are based on the multilingual training on the available corpora $D_B$ and shares the same semantic space for all languages. Therefore, the unified teacher is comprised of different teachers with respective functions simultaneously.

Combining the source-teacher model, the target-teacher model, and the pivot-teacher model, the training objective of our teacher-student training can be described as:
\begin{BigEquation}
\begin{align}
& \mathcal{L}_{S} = \mathcal{L}_{S}^{src} + \mathcal{L}_{S}^{tgt} + \mathcal{L}_{S}^{pivot}
\end{align}
\end{BigEquation}where the parameters of multiple teachers remain unchanged during the training process.

We adopt sequence-level knowledge distillation \cite{syntheticdata_teacher_student} to distill the knowledge from teacher models to the student model in practice. Specifically, we use multiple teacher models to construct the distilled corpora of zero-resource language pairs $D_{S}=\{D_{S_1},\dots,D_{S_M}\}$ with corresponding normalized scores, combined with the original pivot corpora $D_{B}=\{D_{B_1},\dots,D_{B_N}\}$ to train the student model.

As shown in Figure~\ref{overview}, our method can utilize the source-teacher model, target-teacher model, and pivot-teacher model simultaneously to guide the source$\to$target student model, resulting in a more powerful student model.

\subsection{Teacher-Student Transfer}
\label{Training details}

This section will introduce the knowledge distillation details of multilingual multiple teacher-student. Limited by the exponential search space of the source sentences $x$ and $y$, we employ the beam search strategy to generate N-best translation candidates\footnote{Sampling from synthetic data generated by the teacher model according to the probabilities is an easy way to force the student to approximate the teacher \cite{sentence-level-distillation}.} and renormalize the probabilities to teach the student model to approximate the distribution of the teacher model as below:
\begin{BigEquation}
\begin{align}
    w_{x,y} = \frac{exp(w_{x,y}/\tau)}{\sum_{s=1}^{S}exp(w^s_{x,y}/\tau)}
    \label{knowledge-distillation}
\end{align}
\end{BigEquation}where $S$ is the beam size of the fixed teacher model. $w_{x,y}^s$ is the probability of the $s$-th sentence generated by the teacher model. $\tau$ is the temperature. The temperature $\tau \to 0$ increases the weight for the top-selected distilled sentences. We set $\tau < 1.0$ to force the model to focus more on the best-distilled sentence pairs when training.

We first train a single multilingual model with all available pivot corpora $D_B$ as the multiple teacher models for all languages instead of training different bilingual teacher models.

\paragraph{Source-teacher Transfer}
For the source-teacher model, we use the pivot$\to$source model to translate the monolingual pivot sentences of the pivot-target corpus $D_{B}$ into the distilled source sentences. In this way, we get a distilled corpus $D_{S}^{src}$. According to the Equation~\ref{source-teacher-training}, the distilled corpora $D_{S}^{src}$ with the score $w_{x,y}=P(x|z_y;\theta_{z_y \to x})$ is used to teach the student model. 

\paragraph{Target-teacher Transfer}
We adopt the beam search strategies and translate the monolingual pivot language part in the source-pivot corpora $D_{B}$ into target language sentences. Another distilled corpora $D_{S}^{tgt}$ is obtained with the score $w_{x,y}=P(y|z_x;\theta_{z_x \to y})$ for knowledge transfer. 

\paragraph{Pivot-teacher Transfer}
Given the additional monolingual pivot corpora, the pivot sentences are separately translated to the distilled source and target sentences by the pivot-teacher model. We obtain the distilled corpora $D_{S}^{pivot}$ with the score $w_{x,y}=P(x|z_m;\theta_{z_m \to x})P(y|z_m;\theta_{z_m \to y})$ from the monolingual corpus $D_{M}$.

Eventually, with the parameters of the multilingual teacher model fixed, we generate the distilled knowledge and combine them into a whole training dataset $D_{S}=D_{S}^{src} \bigcup D_{S}^{tgt} \bigcup D_{S}^{pivot}$ to train the multilingual source$\to$target student model.

\section{Experiments}
We evaluate our method on the multilingual dataset including 9 languages and 56 zero-resource translation directions. English is the most popular language and there are extensive English-centric data in the real world compared to other languages. Therefore, English (En) is treated as the pivot language in all experiments. 

\subsection{Dataset}
All experiments are conducted on the multilingual dataset of 9 languages extracted from the previous work \cite{zcode}, including English (En), French (Fr), Czech (Cs), German (De), Finnish (Fi), Estonian (Et), Romanian (Ro), Hindi (Hi), and Turkish (Tr).

\paragraph{Bitext Data}  We collect the training data from the latest available year of each language between English and other languages on the WMT benchmark and exclude WikiTiles. The duplicated samples are removed and the number of parallel data of each language pair is limited to 10 million by randomly sampling from the whole corpus. For 72 translation directions of 9 languages, we use the same valid and test sets from TED Talks as the previous work\footnote{\url{ http://phontron.com/data/ted_talks.tar.gz}} for evaluation.

\paragraph{Monolingual Data} The English monolingual data is collected from NewsCrawl\footnote{\url{http://data.statmt.org/news-crawl}} and randomly sample 1 million English sentences. We use a multilingual NMT model to translate these English monolingual data to sentences of other languages as the augmented parallel data, which is utilized as the back-translation data for all baselines.
Our method uses the pivot-teacher model to guide the training of the source$\to$target student model with the monolingual data.

\subsection{Evaluation}
During the inference, the beam search strategy is performed with a beam size of $5$ for the target sentence generation. We set the length penalty as $1.0$. The last $5$ checkpoints are averaged for evaluation. We report the case-sensitive detokenized BLEU using sacreBLEU\footnote{BLEU+case.mixed+lang.\{src\}-\{tgt\}+numrefs.1+smooth.exp+tok.13a+version.1.3.1}.

\subsection{Baselines}
Our method is compared with pivot-based and multilingual baselines. \textbf{Bilingual Pivot-based} \cite{joint_training} translate source to target via pivot language using two single-pair NMT models trained on each pair. \textbf{Multilingual Pivot-based} \cite{multilingual_pivot} leverages a single multilingual NMT
model trained in all available directions for pivot translation. The details of the multilingual baselines are described as follows. \textbf{Multilingual} \cite{multilingual} shares the same vocabulary of all languages and prepends the language symbol to the source sentence to indicate the translation directions. \textbf{Monolingual Adapter} \cite{monolingual_adapter} tunes adapter of each language for zero-shot translation based on a pretrained multilingual model. \textbf{Teacher-Student} \cite{syntheticdata_teacher_student} uses the pivot-target translation model to teach the source-target translation model. \textbf{MTL} \cite{zcode} proposes a multi-task learning (MTL) framework including the translation task and two denoising tasks.

\subsection{Implementation Details}
All experiments are conducted based on the \texttt{Transformer\_big} architecture \cite{transformer}. Both the encoder and decoder contain $6$ layers with $16$ heads per layer. The word embedding size $d_{model}$ is set to $1024$ and the FFN (feed-forward network) size is $4096$. The learning rate is set to $3$e-$4$ with 4000 warmup steps on the multilingual dataset. Adam \cite{adam} is used for updating the parameters. The model with a mini-batch size of $4096$ tokens is trained on 64 Tesla V100 GPUs.

\begin{table*}[t]
\centering
\resizebox{0.85\textwidth}{!}{
\begin{tabular}{l|cccccccc|cc}
\toprule
\toprule
X (High) $\to$ Y (Low) & Fr$\to$Fi & Cs$\to$Fi & Cs$\to$Ro & Cs$\to$Hi & De$\to$Et &Fi$\to$Et & Fi$\to$Ro & Fi$\to$Tr  &Avg$_{8}$ &Avg$_{28}^{>}$ \\
\midrule
\multicolumn{10}{l}{\textit{Train on Parallel Data (Bitext).}} \\
\midrule
Bilingual Pivot \cite{joint_training}                     &13.5 &13.4 &15.2 &2.6 &13.4 &12.7 &13.1 &3.2  &10.9  &9.5  \\
Multilingual Pivot \cite{multilingual_pivot}              &12.5 &11.9 &16.1 &6.9 &14.8 &13.3  &14.0 &5.3 &11.9  &11.2 \\ \midrule
Multilingual \cite{multilingual}                          &3.8  &10.2 &12.6 &5.1 &12.5 &12.0  &10.7 &4.0 &8.9  &8.1 \\
Teacher-Student \cite{syntheticdata_teacher_student}      &13.0 &13.6 &16.4 &7.1 &15.6 &14.6  &14.6 &5.0 &12.5  &10.9  \\
Monolingual Adapter \cite{monolingual_adapter}            &8.2  &10.7 &14.3 &5.9 &12.1 &12.6  &12.4 &4.8 &10.1  &9.2  \\
MTL \cite{zcode}                                          &6.0  &9.0  &13.0 &6.0 &14.3 &12.0  &11.7 &4.6 &9.6  &8.9  \\
\midrule
\textbf{\ourmethod{} w/o pivot-teacher model (our method)} &\textbf{13.8} &\textbf{13.9} &\textbf{16.8} &\textbf{7.3} &\textbf{16.3} &\textbf{14.9} &\textbf{15.1} &\textbf{5.4} &\textbf{12.9} &\textbf{11.8}   \\
\midrule
\multicolumn{10}{l}{\textit{Train on Parallel and Monolingual Data (Bitext + MonoData).}} \\
\midrule
Bilingual Pivot + BT \cite{joint_training}                     &13.9 &13.4 &16.3 &6.9 &15.3 &13.7   &13.6 &4.8 &12.2 &11.0  \\
Multilingual Pivot + BT \cite{multilingual_pivot}              &13.5 &12.6 &16.0 &6.7 &14.8 &13.3   &14.0 &5.6 &12.1 &11.2   \\ \midrule
Multilingual + BT \cite{multilingual}                          &7.5  &10.2 &14.4 &5.7 &12.5 &12.9   &10.7 &5.3 &9.9 &9.4  \\
Teacher-Student + BT \cite{syntheticdata_teacher_student}      &13.6 &13.0 &16.6 &6.8 &15.2 &14.8   &15.2 &5.5 &12.6 &11.6  \\ 
Monolingual Adapter + BT \cite{monolingual_adapter}            &10.8 &7.6  &15.1 &5.0 &15.4 &14.1   &14.1 &5.4 &10.9 &10.0  \\
MTL + BT \cite{zcode}                                          &10.6  &9.0  &13.5 &5.4 &12.7 &12.8   &12.8 &5.2 &10.3  &8.0  \\
\midrule
\textbf{\ourmethod{} (our method)} &\textbf{14.1} &\textbf{14.1} &\textbf{17.1} &\textbf{7.4} &\textbf{16.2} &\textbf{15.0} &\textbf{15.8} &\textbf{5.9} &\textbf{13.2} &\textbf{12.4}  \\ 
\bottomrule
\bottomrule
\end{tabular}}
\caption{X$\to$Y test results for bilingual and multilingual models of 9 language pairs on the WMT benchmark, where the source-pivot corpus is high-resource compared to the low-resource pivot-target corpus. Avg$_{8}$ is the average results of the listed directions and Avg$_{28}^{>}$ is the average BLEU points of all 28 directions under this setting.}
\label{wmt9-high-low}
\end{table*}

\begin{table*}[t]
\centering
\resizebox{0.85\textwidth}{!}{
\begin{tabular}{l|cccccccc|cc}
\toprule
\toprule
X (Low) $\to$ Y (High) & Fi$\to$De & Et$\to$De & Et$\to$Fi &Ro$\to$Cs & Ro$\to$De  & Ro$\to$Et & Tr$\to$Fr & Tr$\to$Et  &Avg$_{8}$ &Avg$_{28}^{<}$ \\
\midrule
\multicolumn{10}{l}{\textit{Train on Parallel Data (Bitext).}} \\
\midrule
Bilingual Pivot \cite{joint_training}                     &15.5 &15.3 &11.0 &14.6 &16.8 &11.8  &10.0 &5.8  &12.6 &11.1  \\
Multilingual Pivot \cite{multilingual_pivot}              &14.6 &16.3 &12.9 &15.1 &18.2 &14.0  &15.7 &9.9  &14.6 &13.6  \\
\midrule
Multilingual \cite{multilingual}                          &11.4 &12.5 &10.1 &12.1 &15.6 &10.7  &7.2  &5.2  &10.6  &9.2 \\
Teacher-Student \cite{syntheticdata_teacher_student}      &16.0 &17.9 &14.1 &16.0 &19.1 &15.1  &16.4 &11.0 &15.7 &13.6  \\
Monolingual Adapter \cite{monolingual_adapter}            &11.8 &14.7 &11.5 &13.1 &16.4 &12.2  &11.7 &7.8  &12.4 &10.4  \\
MTL \cite{zcode}                                          &11.7 &15.1 &10.1 &13.0 &16.1 &12.5  &10.4 &7.0  &12.0 &10.4  \\
\midrule
\textbf{\ourmethod{} w/o pivot-teacher model (our method)} &\textbf{16.6} &\textbf{18.5} &\textbf{14.2} &\textbf{16.3} &\textbf{19.9} &\textbf{15.4} &\textbf{17.1} &\textbf{11.3} &\textbf{16.2} &\textbf{14.7}   \\
\midrule
\multicolumn{10}{l}{\textit{Train on Parallel and Monolingual Data (Bitext + MonoData).}} \\
\midrule
Bilingual Pivot + BT \cite{joint_training}                    &15.0 &17.0 &12.3 &16.0 &18.6 &13.9 &14.6 &9.0  &14.6 &13.8  \\
Multilingual Pivot + BT \cite{multilingual_pivot}             &16.2 &17.4 &12.8 &15.8 &19.4 &14.2 &16.7 &10.4 &15.4 &14.1  \\
\midrule
Multilingual + BT \cite{multilingual}                         &13.6 &16.3 &12.3 &14.9 &16.1 &12.7 &12.1 &8.6  &13.3 &11.3  \\
Teacher-Student + BT \cite{syntheticdata_teacher_student}     &16.6 &19.0 &13.8 &16.5 &20.0 &15.0 &16.8 &10.9 &16.1 &14.3  \\
Monolingual Adapter + BT \cite{monolingual_adapter}           &13.8 &13.8 &11.6 &15.6 &11.7 &13.7 &13.4 &9.6  &12.9 &10.8  \\
MTL + BT \cite{zcode}                                         &12.8 &16.6 &11.5 &13.9 &17.0 &13.0 &14.2 &8.7  &13.5 &11.7   \\
\midrule
\textbf{\ourmethod{} (our method)} &\textbf{17.6} &\textbf{19.6} &\textbf{14.3} &\textbf{17.2} &\textbf{20.7} &\textbf{15.6} &\textbf{17.5} &\textbf{11.5} &\textbf{16.8} &\textbf{15.1}  \\
\bottomrule
\bottomrule
\end{tabular}}
\caption{X$\to$Y test results for bilingual and multilingual models of 9 language pairs on the WMT benchmark, where the source-pivot corpus is low-resource compared to the high-resource pivot-target corpus. Avg$_{8}$ is the average results of the listed directions and Avg$_{28}^{<}$ is the average BLEU points of all 28 directions under this setting.}
\label{wmt9-low-high}
\end{table*}

\subsection{Experiment Results}
The evaluation results over the test sets against the baselines are listed in Table~\ref{wmt9-high-low} and Table~\ref{wmt9-low-high}. The source-pivot corpus is high-resource compared to the low-resource pivot-target corpus in Table~\ref{wmt9-high-low}, and the source-pivot corpus is low-resource compared to the high-resource pivot-target corpus in Table~\ref{wmt9-low-high}.

As observed in Table~\ref{wmt9-high-low}, pivot-based methods including \textbf{Multilingual Pivot} and \textbf{Bilingual Pivot} significantly outperforms the multilingual methods including \textbf{Multilingual}, \textbf{MTL}, and \textbf{Monolingual Adapter}. But pivot-based methods still suffer from error propagation and are computationally expensive by the two-pass translation. On the contrary, our \textbf{\ourmethod{}} method is a unified multilingual source$\to$target model and alleviates this problem. Compared with strong baselines \textbf{Teacher-Student}, our model achieves consistent improvements with $\geq 0.6$ BLEU points gains ($\geq 0.6$ with ``MonoData'') on Avg$_{28}^{>}$. It shows that our student model learns a high-quality representation space across multiple languages with guides of multiple teachers to enforce the capability of zero-resource translation directions. Given the monolingual pivot corpus, our method also beats the pivot-based and multilingual methods by the distilled knowledge from the pivot-teacher model.

In Table~\ref{wmt9-low-high}, the size of the source-pivot corpus is smaller than the pivot-target corpus, so the source teacher transfers more knowledge from the larger pivot-target corpus to guide the student model with more distilled sentences. Our \textbf{\ourmethod{}} also beats all previous methods and gains $\geq 0.9$ BLEU points gains ($\geq 0.5$ with ``MonoData'') on Avg$_{28}^{<}$. It demonstrates the effectiveness and significance of the introduction of source-teacher model. Our \ourmethod{} method without monolingual data (Avg$_{28}^{>}$=$11.8$ and Avg$_{28}^{<}$=$14.7$) performs even better than all baselines with back-translation data.

\subsection{Analysis}

\begin{table}[t]
\centering
\resizebox{0.9\columnwidth}{!}{
\begin{tabular}{ccc|ccc|c}
\toprule
  Source  & Target   & Mono &Fr$\to$De  &De$\to$Ro &Et$\to$Ro & Avg$_{56}$  \\
\midrule
\cmark &        &        &21.3  &17.0  &14.5  &12.3  \\
       & \cmark &        &21.4  &16.2  &15.2  &13.0  \\
       &        & \cmark &22.5  &17.2  &15.4  &12.7  \\
\midrule
       & \cmark & \cmark &22.4  &17.5  &15.8  &13.4  \\
\cmark &        & \cmark &22.3  &16.5  &14.6  &12.6  \\
\cmark & \cmark &        &21.7  &17.5  &15.6  &13.3  \\
\midrule
\cmark & \cmark & \cmark &\textbf{22.8}  &\textbf{17.7} &\textbf{16.4} &\textbf{13.7} \\
\bottomrule
\end{tabular}
}
\caption{Ablation study on different teachers. Avg$_{56}$ denotes the average BLEU points of 56 zero-resource translation directions.}
\label{Source-Target-pivot-teacher}
\end{table}

\paragraph{Effect of Different Teachers} To investigate the effect of the different teachers, we train 7 student guided by all possible combinations of the source-teacher, target-teacher, and pivot-teacher model. Our method combines multiple teachers to direct the source-target student model simultaneously so that our method can improve performance. 
Table~\ref{Source-Target-pivot-teacher} shows ablation results guided by different teachers. Consistently, more teachers can lead to better results, which demonstrates that our proposed model can comprehensively manipulate the advantages of different teachers.

\begin{figure}[t]
\begin{center}
	\includegraphics[width=0.5\columnwidth]{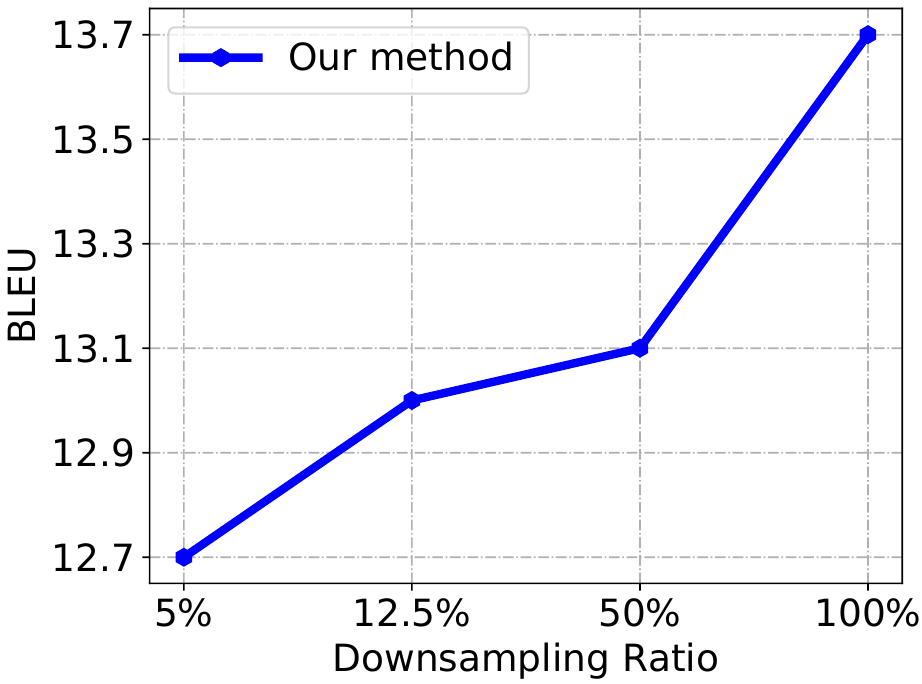}
	\caption{The overall average BLEU points of our method with different downsampling ratios.}
	\label{Size-of-Distilled}
\end{center}
\end{figure}
\paragraph{Size of Distilled Training Data} Given the multilingual pivot corpora of the pivot language and other $N$ languages, $N(N-1)$ distilled training sets of zero-resource directions are used to guide the multilingual student model. The overall scale of corpora contains $TN(N-1)$ sentence pairs, where $T$ is the average size of the distilled corpora. To reduce complexity $\mathcal{O}(TN^2)$ to $\mathcal{O}(TN)$, we adopt a downsampling strategy with $\frac{1}{N}$ downsampling ratio as formulated below:
\begin{BigEquation}
\begin{align}
T' = \max \left\{T_{m}, \, T_{m} + (T - T_{m}) / N \right\}
\label{downsample}
\end{align}
\end{BigEquation}where $T'$ is the size of the downsampled corpus. $T_{m}=1M$ is the threshold to avoid undersampling the low-resource pairs. 

In our work, 16 parallel corpora and 56 ($8 \times 7$) distilled corpora are used for training. The results with different sampling ratios are listed in Figure~\ref{Size-of-Distilled}, which indicates the proper downsampling ratio ($\frac{1}{N}=\frac{1}{8}=12.5\%$) can simultaneously reduce computation cost and get comparable performance. The training time cost of our model is acceptable in a practical scenario. Therefore, our proposed method has been evaluated on the 72 translation directions and can be easily extended to more languages.

\begin{figure}[t]
\begin{center}
	\includegraphics[width=0.5\columnwidth]{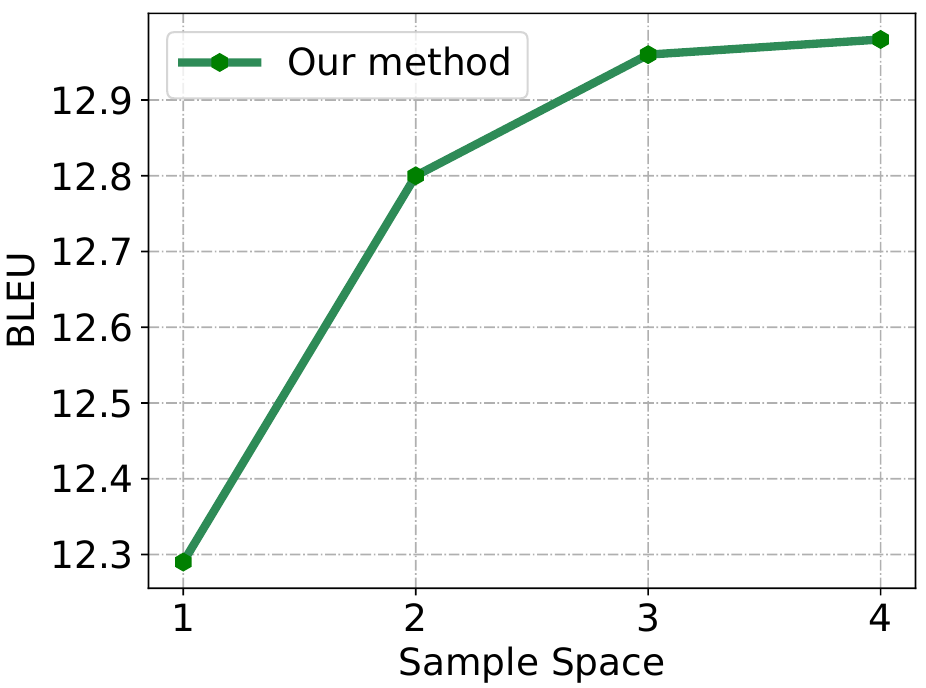}
	\caption{Effect of different beam sizes. We plot the curve of the average BLEU points of all directions with different beams sizes.}
	\label{Sample-Space}
\end{center}
\end{figure}

\paragraph{Sample Space of Possible Sequences} We employ the sequence-level knowledge distillation as formulated in Equation~\ref{knowledge-distillation} and examine our method with different sample space (beam size) settings, where $S$ denotes beam size. Limited by the exponential search space, we use the beam search strategy with a beam size of $S$ ($S \in [1,4]$) to guide the student model. Figure~\ref{Sample-Space} shows that the multilingual student model gains the best performance when $S=3$ or $S=4$ on the zero-resource translation tasks. Considering the computation cost and model performance, we set $S=4$ in our work.

\begin{figure}[t]
\begin{center}
	\includegraphics[width=0.55\columnwidth]{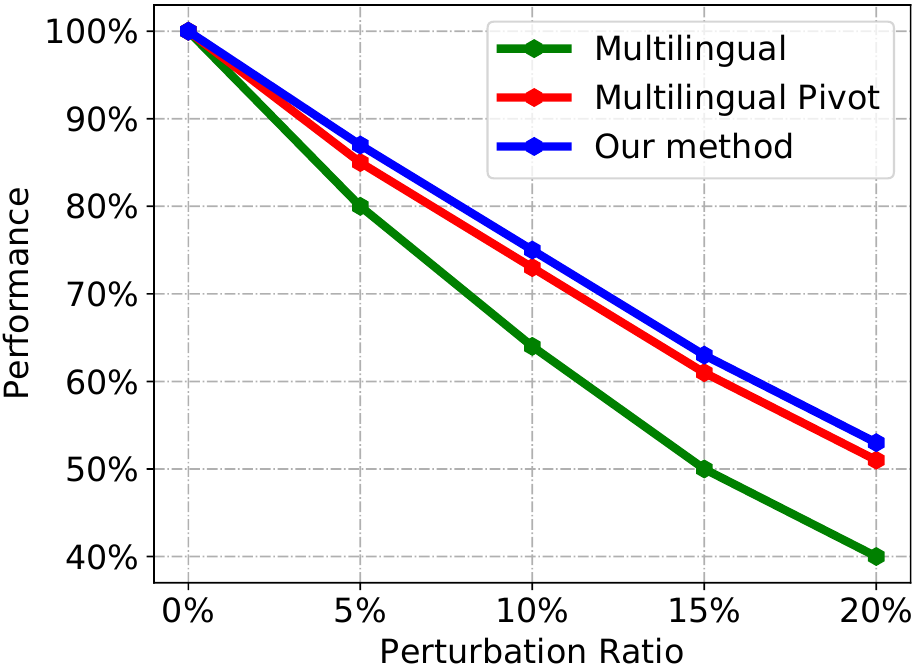}
	\caption{Comparison of different methods on the perturbation experiments with different corrupting probabilities. We display the average performance of all 56 zero-resource translation directions.}
	\label{length}
	\label{Analysis-Error-Propagation}
\end{center}
\end{figure}

\paragraph{Robustness against Input Errors} To further test the robustness of different methods, we add perturbations with different ratios to the source sentence of the test set in Figure~\ref{Analysis-Error-Propagation}. The input sentences are randomly corrupted with four types of perturbations including (1) deletion (drop words), (2) masking (replace words with ``\texttt{[unk]}''), (3) swap (swap words), and (4) substitution (replace words with random words in the vocabulary). For the test set, we randomly perturb source sentences by a fixed corrupting probability.

Given the perturbed input sentence with different corrupting probabilities in Figure~\ref{Analysis-Error-Propagation}, the multilingual model \cite{multilingual} (green line) is easily influenced by the noisy input and degrades to the worst performance. It indicates the multilingual model delivers unstable translation performance for the zero-resource directions that are unseen at training time. The performance of the multilingual pivot-based method \cite{multilingual_pivot} (red line) also consistently drops more than our method due to error propagation introduced by the two-pass translation procedure. The results demonstrate that the multilingual student guided by multiple teachers performs better and avoids error propagation.

\begin{table}[t]
\centering
\resizebox{0.9\columnwidth}{!}{
\begin{tabular}{c|ccc|cc}
\toprule
\#Pairs  &Fr$\to$De &Ro$\to$De &Tr$\to$Cs & Avg$_{16}$ & Avg$_{56}$  \\
\midrule
Supervised  &11.7  &16.1  &9.6  &22.8 &8.7 \\
Zero-resource  &22.0  &19.8  &11.5 &- &13.0 \\
Both  &\textbf{22.8} &\textbf{20.7} &\textbf{12.3} &\textbf{23.1} & \textbf{13.7} \\
\bottomrule
\end{tabular}
}
\caption{Comparison among multilingual student models trained with the original corpora (``Surpervised''), the distilled training corpora (``Zero-resource''), and their combination (``Both'').}
\label{number-of-training-language-pairs}
\end{table}

\paragraph{Number of Training Language Pairs} Our student model is trained on the original parallel corpora $D_{B}$ and the distilled training corpora $D_{S}$ generated by multiple teachers described in Equation \ref{total_training_objective}. The zero-resource translation ability of our student model benefits from the shared semantic space. In Table \ref{number-of-training-language-pairs}, ``Supervised'' denotes the multilingual model trained only with the original corpora of 16 directions, ``Zero-resource'' denotes the student multilingual model trained only with distilled corpora of 56 directions, and ``Both'' denotes our method trained on both the original corpora and distilled corpora. Our \ourmethod{} model trained jointly with 72 directions gets the best performance by transferring the knowledge among different languages.

\section{Related Work}
\paragraph{Zero-Resource NMT} Zero-resource neural machine translation (NMT) is a challenging task since the source-target parallel corpus is not available. A feasible solution is the pivot-based NMT \cite{error_propagation,cascade_5,joint_training,Maximum_Expected,zero_resource_pivot_mono}, where the source language is translated to the pivot language followed by translating the pivot language into the target language. This two-pass translation procedure both increases the complexity and potentially suffers from the error propagation problem because the errors made by the source$\to$pivot model will be introduced to the pivot$\to$target model \cite{multilingual_pivot}. Recent works \cite{syntheticdata_teacher_student,Maximum_Expected,zero_resource_pivot_mono} apply explorations into using the available parallel corpus and the additional monolingual corpus to improve zero-resource performance but limited by the bilingual setting. 
\paragraph{Multilingual NMT} Multilingual neural machine translation (MNMT) \cite{cascade_5,multilingual,multilingual_pivot,bilingual_distillation,unsupervised_MNMT,multilingual_regularization} provides an alternative manner for zero-resource translation without any source-target parallel data but the performance is worse than pivot-based models. The multilingual models with language-aware module \cite{adapter,language_aware_mnmt,monolingual_adapter} are used to translate in zero-resource directions which are unseen at training time. However, the multilingual models often underperform the pivot-based models and deliver poor zero-resource translations. Multilingual pretraining method \cite{pivot-transfer-learning} are used to obtain the crosslingual encoder and then finetunes on the pseudo data. Inspired by previous works \cite{syntheticdata_teacher_student,Maximum_Expected}, we employ multilingual multiple teachers to guide the multilingual source$\to$target student to enhance the zero-resource translation.

\section{Conclusion}
In this paper, we propose a novel method called \textbf{U}nified \textbf{M}ultilingual \textbf{M}ultiple teacher-student \textbf{M}odel for N\textbf{M}T (\textbf{\ourmethod{}}) to ameliorate the translation of zero-resource directions. Our method unifies the source-teacher model, target-teacher model, and pivot-teacher model to guide the multilingual source$\to$target student model, alleviating the error propagation problem caused by two-pass translation. Experimental results on the multilingual dataset of the WMT benchmark corroborate the effectiveness of our method in leveraging the distilled knowledge from the unified teachers.

\bibliographystyle{named}
\bibliography{ijcai22}

\end{document}